\pdfoutput=1

\documentclass[11pt]{article}

\usepackage[preprint]{acl}

\usepackage{times}
\usepackage{latexsym}

\usepackage[T1]{fontenc}
\usepackage[normalem]{ulem} 

\usepackage[utf8]{inputenc}

\usepackage{microtype}

\usepackage{inconsolata}
\usepackage{comment}

\usepackage[table]{xcolor} 

\usepackage{microtype}
\usepackage{tcolorbox}
\usepackage{graphicx}
\usepackage{subcaption}
 \definecolor{lightbrown}{RGB}{210, 180, 140} 
\definecolor{lightgray}{gray}{0.9}
\definecolor{lightblue}{RGB}{173,216,230}
 \usepackage{amsmath} 
\usepackage{algpseudocode}
\usepackage{multirow}
\usepackage{wrapfig}   

\usepackage{inconsolata}
\usepackage{amssymb}
\usepackage{ulem}      

\usepackage{amsmath}  
\usepackage{bm}   
\usepackage{paralist}
\usepackage{enumitem}

\usepackage{colortbl}

\usepackage{boxedminipage}

\usepackage[utf8]{inputenc} 
\usepackage[T1]{fontenc}    
\usepackage{hyperref}       
\usepackage{url}            
\usepackage{booktabs}       
\usepackage{amsfonts}       
\usepackage{nicefrac}       
\usepackage{microtype}      
\usepackage{xcolor}         
\usepackage{enumitem}
\usepackage{makecell}
\usepackage{tabularx}

\title{Learned-Rule-Augmented Large Language Model Evaluators}

\author{Jie Meng\\
  Wuhan University / Wuhan  \\
  \texttt{mengjie@whu.edu.cn} \\\And
  Jin Mao \\
  Wuhan University / Wuhan \\
  \texttt{maojin@whu.edu.cn} \\}

\begin{document}
\maketitle
\begin{abstract}
Large language models (LLMs) are predominantly used as evaluators for natural language generation (NLG) tasks, but their application to broader evaluation scenarios remains limited. In this work, we explore the potential of LLMs as general evaluators across diverse tasks. Although LLM-based evaluators have made progress in different areas, existing methods struggle to generalize due to their reliance on costly, human-designed evaluation principles, which are often misaligned with both annotated data and LLMs’ understanding.To address these challenges, we propose a rule-augmented evaluation paradigm. First, we introduce a rule distillation method that automatically extracts scoring rules from data using an LLM-assisted Monte Carlo Tree Search (MCTS), alleviating scalability issues and improving alignment with data. Second, to enable LLMs to effectively apply the learned rules, we propose two strategies: (1) Chain-of-Rule (CoR), which guides LLM to follow distilled rules, and (2) training a rule-augmented LLM evaluator (RuAE) via reinforcement learning, further bridging the gap between rules and LLMs’ reasoning. Extensive experiments on diverse tasks demonstrate the effectiveness and generalizability of our approach across various evaluation scenarios.
\end{abstract}

\section{Introduction}
Recent advancements in large language models (LLMs) have positioned them as effective and scalable evaluators for assessing generated text quality in Natural Language Generation (NLG) tasks\citep{kocmi2023large,shen2023large}. This raises a natural question: \textbf{can the paradigm of LLMs as evaluators be extended to diverse tasks}?  Studies have explored this potential, investigating LLMs' capabilities in grading essays\citep{mizumoto2023exploring} and assessing citation significance\citep{zhao2025words}. In essence, this means enabling LLMs to quantitatively evaluate text from specific perspectives, for example, assessing quality, measuring expressed tendencies (like empathy or aggressiveness)\citep{wang2024human}, or evaluating textual relationships (like relevance). These applications across multiple domains confirm LLMs' versatility as evaluators. 

Nevertheless, research has revealed challenges in LLM's application as trustworthy general evaluators. Primarily, most existing approaches develop task-specific Chain-of-Thought (CoT) prompts (i.e., evaluation principles)\citep{mizrahi2024state,tornberg2024large}, which are difficult to generalize across diverse tasks. Moreover, these evaluation methods often fail to align with human judgment, manifesting in two key misalignments: 1) \textit{mis-1}:  misalignments between evaluation principles and human-labeled data, and 2) \textit{mis-2}: misalignments between LLMs' understanding and application of these principles. These issues hinder progress toward developing a general text evaluator.

\begin{figure}[h!]
\footnotesize 
    \centering
    {%
\includegraphics[width=0.7\linewidth]{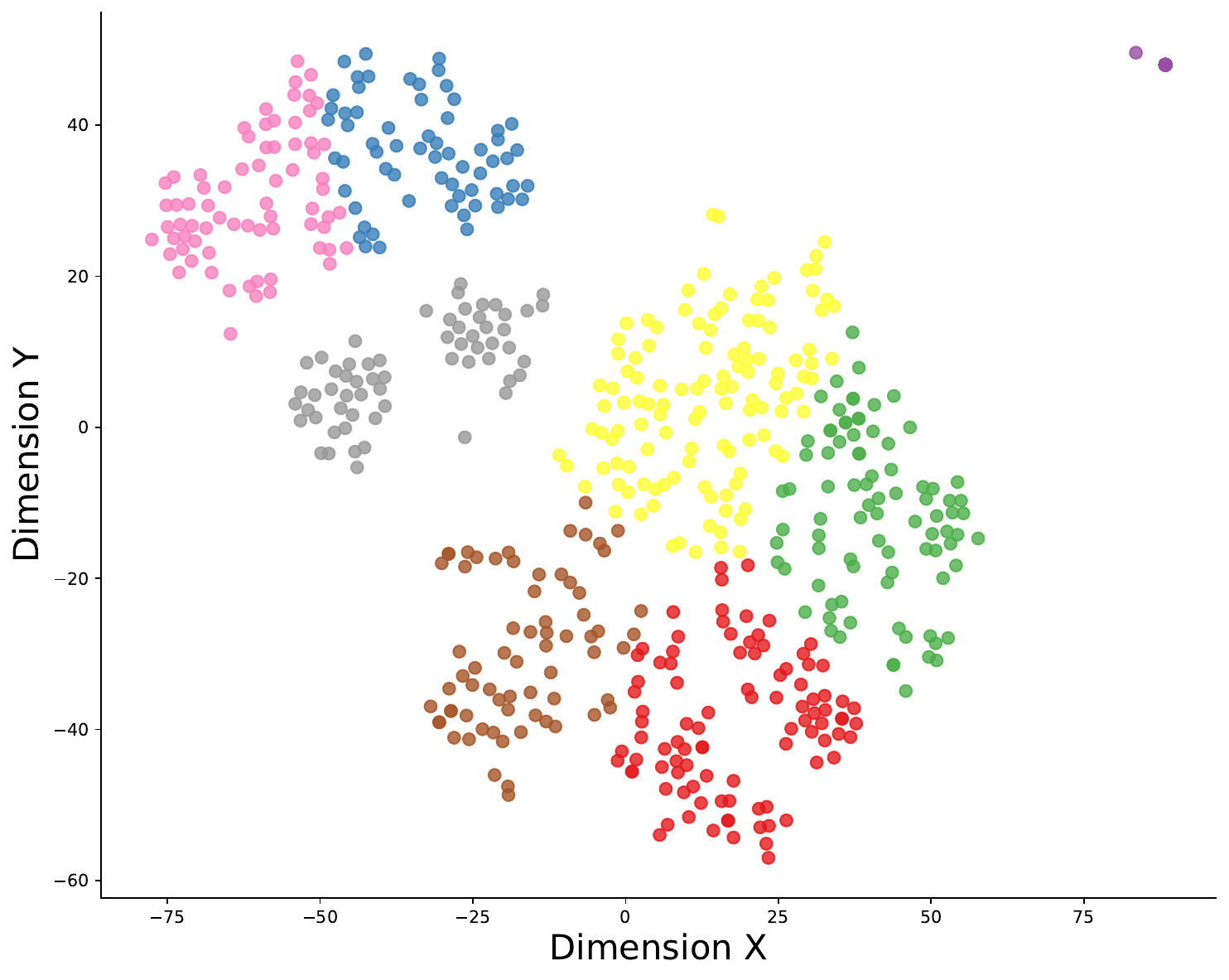}%
    }
    \caption{Clustering visualization of evaluation principles generated by Qwen-7b.}

    \label{fig:image1}
\end{figure}

To investigate the root of these misalignments, we conducted an exploratory study on ASAP (see Section 4.1) by prompting Qwen-7b to propose evaluation principles and score essays accordingly. We analyzed 600 responses, extracted principles and performed dimensionality reduction. As shown in Fig. \ref{fig:image1}, we observed highly dispersed principles with no unified standards. Even within the same evaluation dimension, consistent scoring remained challenging. This contrasts with human evaluation patterns, suggesting that the misalignment primarily stems from differing evaluation standards.

Inspired by this insight, we propose a rule-argumented text evaluation paradigm. These rules represent principles that specify evaluation aspects and detailed criteria for assigning scores. While rules can be manually summarized, this is costly and lacks generalizability. Instead, we focus on learning scoring rules from data. To achieve this, we introduce an LLM-assisted Monte Carlo Tree Search (MCTS)\citep{browne2012survey} approach to distill rules from annotated data, efficiently generating structured and interpretable rules while avoiding compositional search complexity. This approach aligns better with LLMs' understanding and human-labeled data, potentially addressing the misalignment \textit{mis-1}.\\

When attempting to enable LLMs to apply these rules, we encounter another misalignment challenge \textit{mis-2}: how can LLMs effectively follow rules during evaluation? Even well-crafted rules cannot prevent deviations in LLMs' execution\citep{calderon2025alternative}, whether in providing rationales or assigning scores. To address this, we explore two strategies to enhance LLMs' reasoning with learned rules:
1) Chain-of-Rule (\textbf{CoR}): injecting distilled rules into prompts, a simple and scalable method. However, it may not fully resolve misalignments, especially if the LLMs for rule distillation and application differ.
2) Training a rule-augmented LLM evaluator (\textbf{RuAE}). While supervised fine-tuning (SFT) \cite{Trivedi2024SelfrationalizationIL}  is a typical choice for training LLMs, its limitations include overfitting, catastrophic forgetting, and and lack of alignment with human judgment due to limited reasoning process data. 
 Inspired by recent advancements in reasoning models, \cite{guo2025deepseek}. we train an LLM evaluator using reinforcement learning (RL). We design a composite reward function and train the model using Group Relative Policy Optimization \citep{shao2024deepseekmath}.
We conduct extensive experiments on four diverse tasks (grading, regression, ranking, and judging). Results show that CoR improves performance across models of different sizes, while RuAE outperforms larger reasoning models on complex tasks. Moreover, the learned rules are reasonable and human-algined.  The results demonstrate the effectiveness and generalizability of our approach.
Our contributions are as follows:
\begin{itemize}
  \setlength\itemsep{0pt}      
  \setlength\parskip{0pt}      
  \setlength\parsep{0pt}       
    \item We propose a scalable method to learn interpretable scoring rules from data.
    \item We introduce a rule-augmented LLM evaluator trained with reinforcement learning, improving reasoning ability in evaluation.
    \item Extensive experiments demonstrate the effectiveness of our approach across various scenarios, extending the domain coverage of LLM-based evaluators. Results further highlight the importance of enhanced reasoning ability, especially in tasks involving complex criteria and long-form text.
\end{itemize}

\section{Related Work}
\textbf{LLM as evaluators}. LLM evaluators have emerged as cost-effective alternatives to human assessment in multiple generative tasks including machine translation\citep{kocmi2023large}, summarization\citep{shen2023large}, and dialogue generation\citep{liu2024measuring}. By equipping models with evaluation criteria and instructing them to generate scores to assess the quality of candidate text, score-based LLM evaluators have been developed that can be easily adapted to various tasks through instruction tuning \citep{mizrahi2024state}. Automated Essay Scoring represents one of the earliest tasks to explore this capability\citep{chen2013automated}. Initial approaches utilized zero-shot evaluator prompts with GPT-2 to return overall scores based solely on given scoring criteria\citep{mizumoto2023exploring}. Yancey et al. revealed the advantages of LLMs by referencing detailed scoring rubrics or generating reasoning before scoring\citep{yancey2023rating}. LLM evaluators have since been applied to additional domains, such as automatic annotation in computational social science for measuring political leanings\citep{tornberg2024large}, as well as citation importance evaluation in literature mining\citep{huang2023citation}. Nevertheless, these methods remain difficult to scale due to their reliance on domain-specific expert prompt engineering.\\

\noindent\textbf{Reasoning with LLMs}. Early work has primarily proposed Chain-of-Thought (CoT) to encourage step-by-step thinking \citep{wei2023chainofthoughtpromptingelicitsreasoning}. Building on it, non-linear thinking approaches utilizing tree or graph structures are proposed to guide CoT generation\citep{yao2023tree}. Drawing parallels to tree reasoning, some studies explore Monte Carlo Tree Search (MCTS) to enhance LLM \cite{xie2024montecarlotreesearch}. Several methods attempt to optimize LLMs by leveraging reasoning paths identified through MCTS. However, a significant challenge with MCTS is that token-level states create an extremely large search space, resulting in inefficient search. In contrast, our MCTS approach searches at rule-level, substantially reducing complexity.
Beyond search strategies, Reinforcement Learning (RL) offers another pathway to improve the intrinsic reasoning ability of language models\cite{pang2023languagemodelselfimprovementreinforcement,feng2024improvinglanguagemodelreasoning}. PPO and DPO algorithms were initially applied to align models with human by training a reward model from pairwise preferences\cite{christiano2023deepreinforcementlearninghuman, stephan2023trimmingestimatorlatentdiffusionobservedadoptionmodel}. Recently, to efficiently apply reinforcement learning for optimizing reasoning models, DeepSeek proposed the GRPO algorithm \cite{shao2024deepseekmath}, eliminates the need for a critic model by estimating baselines from group scores, enabling efficient policy model training. Similarly, RLOO\cite{ahmadian2024basicsrevisitingreinforcestyle} introduces a simplified REINFORCE-style \cite{williams1992simple} 
 optimization framework, has shown improvements in pure RL training.

\begin{figure*}[h]
    \centering
    \includegraphics[width=1\textwidth]{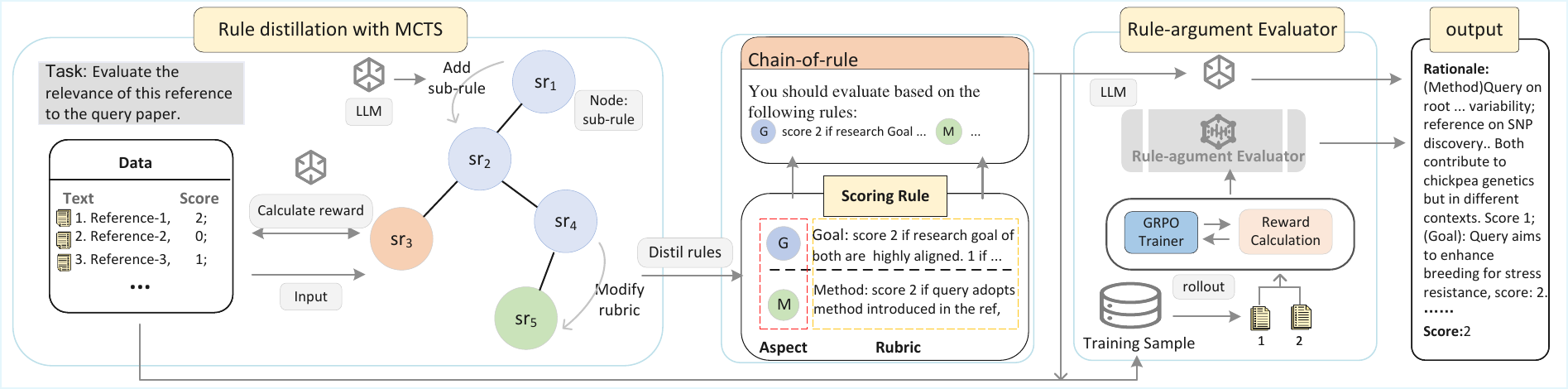} 
    \caption{Overview of Our Method}
    \label{fig:method}
\end{figure*}

\section{Method}
We formalize the process of LLM evaluators as \( p_{\mathcal{M}}(s, RT\mid y;  I) \), where \( y \) is the input text, \( I \) is the instruction, \( s \) is the score, and \( RT \) is the rationale. Existing CoT-like methods introduce additional human-designed guidance (\( G \)), extending the formulation to \( p_{\mathcal{M}}(s, RT \mid y; I, G) \). However, such guidance may not always align with data. For simplicity, we omit the \( RT \) and \( I \) terms hereafter. We decompose the scoring process as, $p_{\mathcal{M}}(s \mid y) \propto p_{\mathcal{M}}(y \mid s) \times p_{\mathcal{M}}(s)$.
Here, \( p_{\mathcal{M}}(y\mid s) \) captures what a text with score \( s \) should look like. Our key insight is to learn interpretable scoring rules (\(\mathcal{R} \)) from data, so that \( p_{\mathcal{M}}(y\mid s) \) better approximates human judgment \( p_{\mathcal{H}}(y\mid s) \).
\begin{equation}
   \mathcal{R}^* = \arg\min_{\mathcal{R}} \; \underset{s \sim \mathcal{D}}{\mathbb{E}} \left[ \mathcal{L}\left(p_{\mathcal{M}}(y \mid s;\mathcal{R}),\; p_{\mathcal{H}}(y \mid s)\right) \right] 
\end{equation}
Based on this idea, our method is structured in two stages. As shown in Fig. \ref{fig:method}, the first stage is to distill interpretable scoring rules from data. Then, we apply these rules to guide the evaluation process in two ways: (1) by prompting LLMs with the learned rules (\textbf{CoR}), and (2) by further aligning the evaluator to these rules and prior $p_{\mathcal{M}}(s; I)$ via reinforcement learning (\textbf{RuAE}), ensuring both the rationale $RT$ and the scoring are consistent with human and rules.
\subsection{Scoring Rule Distillation with Mento Carlo Tree Search}
\textbf{Scoring Rule}. A scoring rule $\mathcal{R}$ is formalized as a set of sub-rules $\{sr_1,sr_2,...sr_k\}$, where each sub-rule encapsulates a specific evaluation criterion. Formally, a sub-rule defined as a tuple $sr = (\mathrm{asp}, \mathrm{ru})$, where $\mathrm{asp}$ represents the aspect of evaluation (e.g., clarity or coherence in NLG evaluation), and is required to be independent from other aspects; $\mathrm{ru}$ denotes the rubric, a detailed scoring guideline for the corresponding aspect.

\noindent \textbf{States, Actions in MCTS}. With the scoring rule defined, the state, action, and reward in MCTS for scoring rule searching can be defined as:
\begin{itemize}[topsep=0pt, partopsep=0pt, itemsep=0pt, parsep=0pt,left=1em]
\item \textbf{State} ($\mathcal{S}$): Each state represents a candidate scoring rule, consisting of a set of sub-rules in the search process, $S_n = \{sr_1,sr_2,...sr_k\}$. The initial state is the empty set, $S_0 = \emptyset$. 
\item \textbf{Action} ($\mathcal{A}$): Actions are categorized into two types: (i). Adding a new sub-rule to the current state, i.e., $\mathcal{A}_{add}: S \rightarrow S \cup sr_{i}$, where $sr_{i}$ is generated by LLM. (ii). Modify the rubric of an existing sub-rule in the current state,i.e., $\mathcal{A}_{mod}: S \{sr_i(\mathrm{asp}, \mathrm{ru})\} \rightarrow S \{sr_i(\mathrm{asp}, \mathrm{ru'})\}$, where rubric $\mathrm{ru'}$ is modified by LLM.
\end{itemize}

\noindent \textbf{Sub-rule Generation}. 
Actions in the action space are generated by LLMs. For $\mathcal{A}_{add}$, a candidate pool of sub-rules can be pre-generated before the search to ensure independence among sub-rules while maintaining action space stability. When generating new sub-rules, specific aspects or domain-specific constraints can be specified to reduce ineffective exploration if some evaluation dimensions are known. For action type $\mathcal{A}_{mod}$, despite numerous possible modification directions, we restrict modifications to two directions: stricter or more lenient rubrics. Prompts are provided in Appendix \ref{appendix:Prompt}. 

\noindent \textbf{Simulation}. 
This stage calculates the reward of a new node to evaluate the rule's quality after changing a sub-rule. We use a dataset $D$ and an LLM as the environment. The rule's effectiveness is measured by comparing LLM-predicted scores $\hat{y}_i = \mathbf{LM}(d_i|S)$ to the ground truth $y_i$,  
 $\mathbb{E}_{d_i \sim D} \left[ \mathcal{C}(\hat{y}_i, y_i) \right]$. Here, $\mathcal{C}(\cdot, \cdot)$ is a task-specific measure, such as mean squared error in regression.
Evaluating each new state across the dataset is computationally costly. For efficiency, we assume independent sub-rules have equal weight, calculating the total predicted score as: $\hat{y}_i = \frac{1}{k} \sum_{j=1}^kp_j$,
where $p_k = \mathbf{LM}(d_i|sr_k)$ represents the prediction from the $k$-th sub-rule, and a caching buffer stores prior predicted scores $(p_1, p_2, \dots, p_{k-1})$. This approach can also minimizes context window, ensuring independent aspect assessment.

\noindent \textbf{Selection and Expansion}. The selection phase starts from the root node and iteratively chooses the most promising child node using the Upper Confidence Bound applied to Trees (UCT) formula, until a leaf node is reached. \begin{equation}
    \text{UCT}_i = \frac{W_i}{N_i} + C \cdot \sqrt{\frac{2 \ln N_p}{N_i}},
\end{equation}

where $W_i$ is node $i$'s total reward, $N_i$ is its visit count, $N_p$ is the parent's visit count, and $C$ adjusts exploration. If this leaf node is not terminal, the expansion phase creates new child nodes for unexplored actions. 

To enhance search efficiency and explore novel evaluation aspects, our algorithm divides search process into two stages. The initial expansion \textit{stage-1} limits the action space to $\mathcal{A}_{add}$ for adding sub-rules. When the state with the maximum number of sub-rules in the search tree reaches a predefined threshold, the process transitions to the \textit{stage-2}, where the action space switches to $\mathcal{A}_{mod}$ for refining existing sub-rules. To balance exploration and exploitation, the UCT exploration coefficient $C$ is higher in \textit{stage-1} for exploration and lower in the \textit{stage-2} for exploitation.

\noindent  \textbf{Backpropagation}. The reward obtained from the simulation is propagated upward through the selected path from the leaf node to the root. 

\subsection{Rule-Augmented Evaluator}
At this stage, we utilize the rules learned from the MCTS phase to enhance the LLM's reasoning and scoring abilities as a evaluator. We introduce two approaches: Chain-of-Rule (\textbf{CoR}) and Rule-Augmented Evaluator (\textbf{RuAE}).

\subsubsection{Chain-of-Rule} We prepend the learned rules to the task instructions to guide LLMs in performing rule-based evaluations, enabling structured assessment without training. Follow the steps below for CoR: 1) Rule Filtration: Filter out rules with extremely low frequency to ensure that only meaningful  rules are considered, eliminating noise. 2)  Selection: Calculate the average reward of each rule and select the top 5 as scoring rules, constitutes high-value rules. 3) Generation: Randomly sample a rule from this set, inserting after the task instruction and the text to be evaluated to form a prompt. The prompt instructs the LLM to follow the rule and evaluate the text from multiple aspects. Details in \ref{appendix:Prompt}.

\subsubsection{Train Rule-Augmented Evaluator with Reinforcement Learning}
\textbf{Rule-guided Prompt Construction}. We adopt a prompt scheme similar to that in Chain-of-Rule, but with modifications tailored for comparative evaluation. Each prompt evaluates a pair of text samples $(t_1,t_2)$ for comparison. To maintain balance between exploration during rollout, we expand the ruleset by incorporating sub-rules from high-value rules in CoR, keeping only the highest-reward version when aspects overlap. The instruction also prompts the LLM to select key aspects from these sub-rules and assign weights accordingly. Detailed implementation is provided in the Appendix \ref{appendix:details}.

\noindent\textbf{Reward Design}. To ensure the LLM follows rules while producing reasonable scores, the reward design addresses two key objectives: 1) the predicted scores should closely approximate the ground-truth values, and 2) the model should accurately distinguish the relative differences between texts. Therefore, we design a reward function that combines both absolute score accuracy and relative ranking preservation when comparing the predicted scores $(s_1, s_2)$ against the ground-truth labels $(s_1^*, s_2^*)$. The reward consists of two components with equal weights:
\begin{align}
r_{\text{order}} &= 
\begin{cases}
1, & \text{if } \text{sgn}(s_1 - s_2) = \text{sgn}(s_1^* - s_2^*) \label{eq:rorder}\\
-1, & \text{otherwise}
\end{cases} \\
r_{\text{abs}} &= \sum_{i=1,2} \{1 - 2 \cdot \frac{|s_i - s_i^*|}{sc}\}
\end{align}
where $sc$ represents the score range.

\noindent\textbf{RL Training}. 
We formulate the RL objective function as follows:
\begin{equation}
\begin{aligned}
    \max_{\pi_\theta} \; \mathbb{E}_{x \sim \mathcal{D},\, y \sim \pi\theta(\cdot \mid x)} \left[ r_\phi(x, y) \right] -   \\
    \beta D_{\mathrm{KL}}\left[ \pi_\theta(y \mid x) \,\|\, \pi_{\mathrm{ref}}(y \mid x) \right]
\end{aligned}
\end{equation}
where where $\pi_\theta$ is the policy LLM, $\pi_{\mathrm{ref}}$ is the reference LLM, $ r_\phi$ is the reward function and $D_\text{KL}$ represents the KL divergence.  To improve policy optimization stability and avoid the need for an additional value function approximation, we use Group Relative Policy Optimization (GRPO)\cite{shao2024deepseekmath}, which leverages the average reward of multiple sampled outputs as a baseline.

\section{Experiment}
\subsection{Experimental Setup}
\textbf{Datasets and evaluation metrics. } Our method is designed to adapt to various evaluation and scoring tasks. To validate its effectiveness, we conduct experiments on diverse types of tasks, including scoring, regression, ranking. \\
\textit{1). \textbf{ASAP}}\citep{ramesh2022automated}: A benchmark dataset for automated essay scoring (AES). The evaluation metrics are Quadratic Weighted Kappa (QWK) and Kendall-Tau correlation ($\tau$). 
  \textit{2). \textbf{Relish}}\citep{relish2019}: A scientific document recommendation task, which is requried to evaluate the relevance of candidate papers to a query. It represents proximity scoring. Metrics include mean average precision (mAP), and normalized discounted cumulative gain (nDCG).
\textit{3). \textbf{Amazon}} \citep{ni2019amzon}: A dataset for predicting 5-star ratings of product reviews, representing ordinal regression tasks. Evaluation metrics include mean absolute error (MAE) and mean squared error (MSE). \textit{4). \textbf{SummEval}}\citep{fabbri2021summeval}: A meta-evaluation benchmark for machine-generated summarization. Spearman ($\rho$) and Kendall-Tau ($\tau$) correlation are used to measure the agreement between ground truth and predictions. 

\noindent \textbf{Baselines}. We compare our proposed methods against four categories of baselines:\\
(1) \textit{Base models}: We select three open-source base models with different parameter scales: Deepseek-v2.5 (236B)\cite{liu2024deepseek}, Qwen-32B, and Qwen-7B\cite{bai2023qwen}. For these models, we evaluate two approaches: \textit{Vanilla Scoring}, which directly prompt models without additional guidance, and CoT-like Scoring, which guides LLM first generates scoring criteria before evaluation. The known scoring dimensions are included in the prompts if available.  
(2) \textit{Large Reasoning Models}: We include two advanced reasoning models: Deepseek-R1 (671B)\cite{guo2025deepseek} and QwQ (32B) \cite{yang2025qwen3}.
(3) \textit{Task-Specific Baselines}: For non-LLM-based methods, the task-specific models are shown in \ref{tab:tasks}.
(4)  \textit{SFT}. Perform full-parameter fine-tuning on Qwen-7B using the same training data.

\begin{table}[ht]
\centering
\caption{Task-specific Baselines}
\label{tab:tasks}
\small
\begin{tabular}{ccc}
\toprule
\textbf{Task} & \textbf{Baseline-1 (M-1)} & \textbf{Baseline-2 (M-2)} \\
\midrule
\textbf{ASAP}     & \makecell{BERT \\ \scriptsize\citep{devlin2019bertpretrainingdeepbidirectional}} 
                  & \makecell{RoBERTa \\ \scriptsize\citep{liu2019robertarobustlyoptimizedbert}} \\
\midrule
\textbf{Relish}   & \makecell{SciBERT \\ \scriptsize\citep{beltagy2019scibertpretrainedlanguagemodel}}
                  & \makecell{SPECTER \\ \scriptsize\citep{cohan2020specterdocumentlevelrepresentationlearning}} \\
\midrule
\textbf{Amazon}   & BERT
                  & RoBERTa \\
\midrule
\textbf{SummEval} & \makecell{UniEval \\ \scriptsize\citep{zhong-etal-2022-towards}}
                  & \makecell{BARTScore \\ \scriptsize\citep{yuan2021bartscoreevaluatinggeneratedtext}} \\
\bottomrule
\end{tabular}
\end{table}

\noindent\textbf{Implementation Details}. In rule distillation phase, we utilize DeepSeek-v2.5 to assist in distillation, which includes rule generation, rule modification, and score prediction. RuAE training is based on Qwen2.5-7B-Instruct. The training is conducted over a single epoch, with a maximum prompt length of 2048 tokens. We trained it using 5×A800 80GB GPUs. Details can be found in Appendix \ref{appendix:details}. 

\subsection{Overall Results}

\begin{table*}[h!]
\footnotesize
\centering
\begin{tabular}{lcccccccc}
\toprule
\multirow{1}{*}{Method} & \multicolumn{2}{c}{ASAP} & \multicolumn{2}{c}{Relish} & \multicolumn{2}{c}{Amazon} & \multicolumn{2}{c}{SummEval} \\
\cmidrule(lr){2-3} \cmidrule(lr){4-5} \cmidrule(lr){6-7} \cmidrule(lr){8-9}
 & QWK & $\tau$  & mAP & nDCG & MAE & MSE & $\rho$ & $\tau$ \\ 
\midrule

\multicolumn{9}{c}{D\textsc{eep}S\textsc{eek}} \\ 
\midrule
Scoring   & 0.086 & 0.124 & 0.231 & 0.770 & 1.182 & 2.184 & 0.009 & 0.010 \\
CoT   & 0.0553 & 0.174 & 0.256 & 0.791 & 0.312 & 0.387 & 0.493 & 0.401 \\
DeepS\textsc{eek}\textsubscript{-R1}  & \textcolor{gray}{\uline{\textbf{0.315}}} & 0.279 & 0.210 & 0.799 & 0.222 & 0.246 & 0.521 & 0.432 \\
\rowcolor{gray!20} CoR  & 0.298 & \textcolor{gray}{\uline{\textbf{0.323}}} & \textbf{0.320} & \textcolor{gray}{\uline{\textbf{0.826}}}
 & \textbf{0.209} & \textbf{0.234} & \textcolor{gray}{\uline{\textbf{0.586}}} & \textcolor{gray}{\uline{\textbf{0.467}}} \\

\midrule
\multicolumn{9}{c}{Qwen-2.5 \textsubscript{-$32$B} } \\ 
\midrule
Scoring   & 0.131 & 0.235 & 0.294 & 0.818 & 1.212 & 2.235 & 0.546 & 0.453 \\
CoT   & 0.071 & 0.223 & 0.287 & 0.829 & 1.119 & 2.004 & 0.569 & 0.477 \\
QWQ\textsubscript{-$32$B}   & 0.106 & 0.207 & 0.285 & 0.806 & \textcolor{gray}{\uline{\textbf{0.215}}} & \textcolor{gray}{\uline{\textbf{0.327}}}
 & 0.190 & 0.151 \\
\rowcolor{gray!20} CoR & \textcolor{gray}{\uline{\textbf{0.163}}} & \textcolor{gray}{\uline{\textbf{0.258}}} & \textcolor{gray}{\uline{\textbf{0.319}}} & \textcolor{gray}{\uline{\textbf{0.849}}}
& 1.123 & 2.017 & \textbf{0.600} & \textbf{0.506} \\

\midrule
\multicolumn{9}{c}{Qwen-2.5 \textsubscript{-$7$B} } \\ 
\midrule
Scoring   & 0.286 & 0.298 & 0.291 & 0.821 & 1.21 &  2.28 & 0.401 & 0.335 \\
CoT & 0.122 & 0.231 & \textcolor{gray}{\uline{0.293}} & 0.824 & 1.18 & 2.21 & 0.398 & 0.329 \\ 
\rowcolor{gray!20} CoR & \textcolor{gray}{\uline{0.316}}  & \textcolor{gray}{\uline{0.325}} & \textcolor{gray}{\uline{0.293}} &  0.826 & 1.20 & 2.30 & \textcolor{gray}{\uline{0.456}} & \textcolor{gray}{\uline{0.383}} \\

\rowcolor{blue!20} RuAE\textsubscript{-$7$B} & \textbf{0.379} &  \textbf{0.326} & \textcolor{gray}{\uline{\textbf{0.301}}}
 &  \textbf{0.934} &  \textcolor{gray}{\uline{\textbf{0.366}}}  & \textcolor{gray}{\uline{\textbf{0.519}}} & \textcolor{gray}{\uline{\textbf{0.466}}} &  \textcolor{gray}{\uline{\textbf{0.404}}}
 \\
 SFT\textsubscript{-$7$B} & 0.017 & 0.089 & 0.234 & \textcolor{gray}{\uline{0.874}} & 1.93 & 5.54 & 0.168 & 0.132 \\
\midrule
\multicolumn{9}{c}{Task-Specific Baselines} \\ 
\midrule
M-1  &  0.017 & -0.027 & 0.109 & 0.853 & 1.22 & 2.41 & 0.364 & 0.323 \\
M-2  & 0.0021 & -0.001 & 0.241 & 0.876 & 0.226 & 0.235 & 0.435 & 0.403 \\

\bottomrule
\end{tabular}
\caption{Overall Performance. The best for the same-parameter models is marked with gray \textcolor{gray}{\uline{\textbf{bold}}}, while the best across all models is marked with \textbf{bold}. The second-best score for 7B model is marked with underlined \textcolor{gray}{\uline{bold}}.}
\label{tab:Main results}
\end{table*}
\textbf{Performance on different (families of) LLMs}. As shown in \ref{tab:Main results}, when comparing training-free methods, CoR demonstrates consistent performance improvements in most tasks. For instance,
on SummEval, CoR delivers substantially superior performance compared to other methods, even surpassing models with larger parameter scales. On Relish, CoR surpasses DeepSEEK-R1 by over 52\% in mAP. As model size decreases, CoR's advantage becomes even more pronounced on the ASAP, where it significantly outperforms CoT and Scoring for smaller models like 7B. This suggests that CoR is especially effective for tasks requiring complex reasoning, where smaller models often struggle. However, CoR's performance on the Amazon is inconsistent. It slightly outperforms DeepSeek-R1 but shows limited improvement across other parameter sizes. This may be due to the task's nature, short text analysis, which smaller models can manage effectively without advanced reasoning. Overall, CoR consistently outperforms Scoring and CoT, highlighting the effectiveness of learned rules to guide LLMs in complex analysis.\\
\noindent \textbf{Comparison across all models}. Across all parameter sizes, RuAE achieves the best performance on tasks ASAP and Relish,surpassing even the largest model, DeepSeeK-R1. On ASAP's QWK, RuAE leads by 20.3\% over the second-best model (R1), and on Relish's nDCG, it exceeds CoR by 10\%. This demonstrates that RL-trained models like RuAE can effectively align with scoring rules and deliver highly accurate results, particularly for professional, long-text tasks such as literature mining and essay scoring. On the other hand, RuAE's performance on Amazon and SummEval is less dominant. While RuAE achieves the best results among models of comparable sizes, it does not outperform QWQ on Amazon. This may stem from the lower reasoning demands of short-text tasks or differences in the base models. Notably, the Qwen-32B family achieves the highest SummEval scores across all methods, suggesting that its training data may include task-specific optimizations.

\subsection{Ablation Studies}
\textbf{Framework Ablation}. To comprehensively evaluate the impact of various components on the performance of our method, we conducted an ablation study by modifying key components of our framework. Specifically, we focus on the following experiment:
1) \textbf{Reward Ablation (sg-rw)}: we removed the \( r_{order} \) component from the reward function, retaining only \( r_{abs} \), to assess the effectiveness of the composite reward design.
2) \textbf{MCTS+SFT}: To investigate the role of reinforcement learning in training, we utilized MCTS-generated trajectories for fine-tuning. Specifically, we collected high-reward states along with their corresponding LLM reasoning processes, formatted them into question-answer pairs and used them to fine-tune the Qwen-7B model.
3) \textbf{MCTS Only}: we also compared the effect of rule-based prompting without model training, which is our CoR for Qwen-7B, to validate the standalone impact of learned rules.
\begin{figure}[h]
    \includegraphics[width=0.47\textwidth]{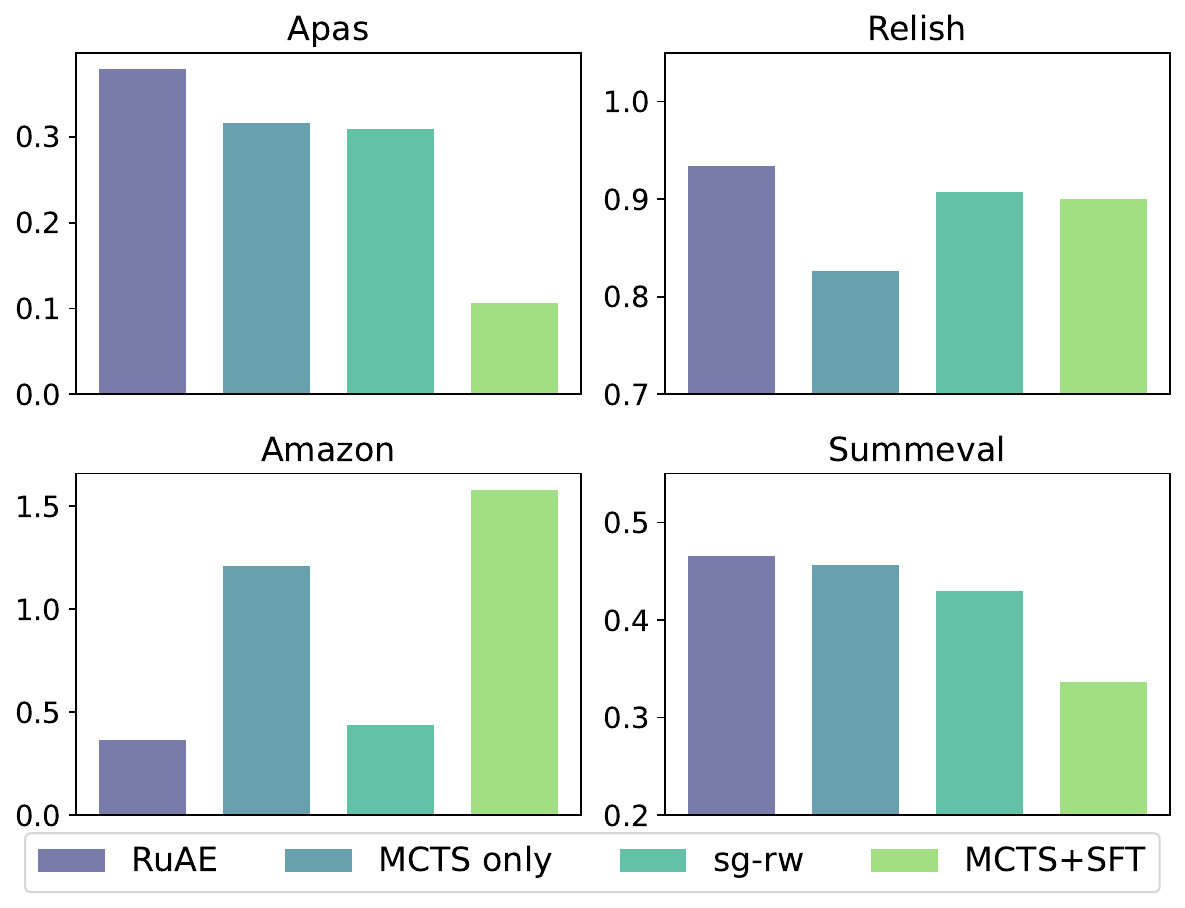} 
    \caption{Framework ablation study. We select these metrics for comparision: QWK(ASAP), nDCG(relish), MAE(Amazon), $\rho$(SummEval)}
    \label{fig:ablation}
\end{figure}

The results in Figure \ref{fig:ablation} indicate that the variants result in performance reduction, underscoring the importance of each component in our method. Among these, the MCTS+SFT exhibited the most significant drop in performance, as analysis revealed that high-reward data often focuses on easily evaluable samples, introducing bias. Moreover, the training samples may have inaccurate scoring in specific dimensions even the result is correct. The reward ablation (sg-rw) had a minimal impact on the Amazon but a noticeable decline on ASAP, which prioritizes ordinal relationships. Conversely, Amazon data shows less inter-text dependency, highlighting that \( r_{order} \) plays a crucial role in learning ordinal relationships. MCTS only, which relies on prompting, showed relatively stable performance but experienced declines on the Amazon and Relish. This indicates that the prompting approach may struggle to align with score distributions due to the absence of training.

\noindent\textbf{Ablation on Rule Distillation }.
We conducted an ablation study to investigate the impact of different reward computation approaches on rule distillation, as they directly affect the selection of rules. Specifically, we compared our standard method, which uses overall dataset-level metrics, with an alternative approach named \textbf{pairwise reward} \textit{(PAR)}. The PAR adopts the same reward calculation formula as described in Section 3.2. To assess which method produces more stable and unified rules, we analyzed the top-10 rules from each approach. We calculated their information entropy ($\mathcal{H}$) and the average Jaccard similarity coefficient ($JS$) (by averaging all pairwise $JS_{ij}$ values among the rules). 

\begin{figure}[htbp]
    \centering
    \begin{subfigure}[b]{0.57\textwidth}
        \centering
        \includegraphics[width=\textwidth]{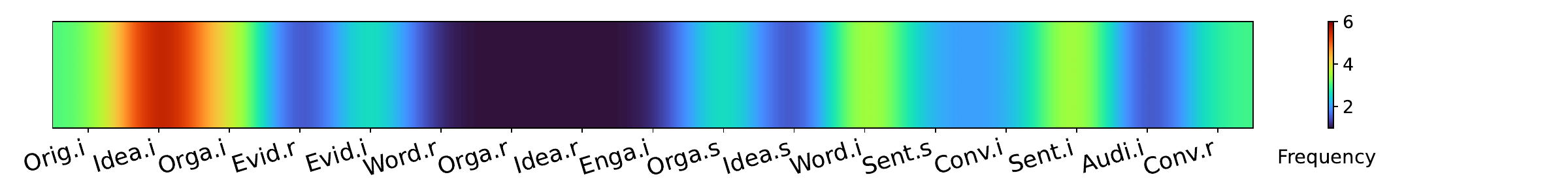}
       \caption{PAR}
        \label{fig:sub1}
    \end{subfigure}
    \hspace{0.03\textwidth} 
    \begin{subfigure}[b]{0.57\textwidth}
        \centering
        \includegraphics[width=\textwidth]{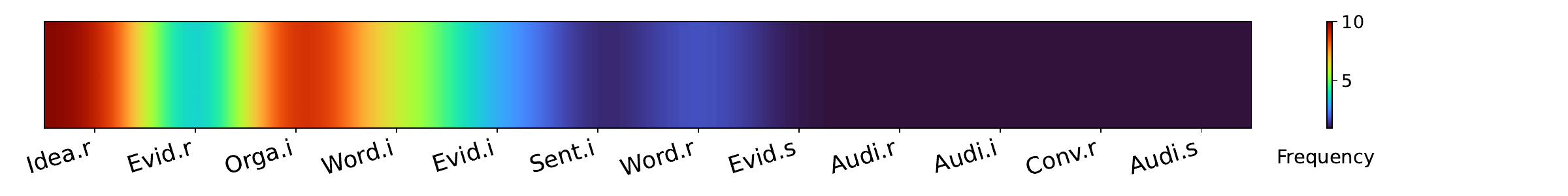}
         \caption{Ours}
        \label{fig:sub2}
    \end{subfigure}
 \caption{Spectrogram of sub-rule frequency comparison between Ours and PAR}
   \label{fig:PAR}
\end{figure}

Results show that PAR yields rules with higher $\mathcal{H}$ and lower $JS$,indicating greater diversity but less consistency and stability. This suggests that PAR struggles to discover a set of clear and uniform rules for scoring.
\begin{wraptable}{r}{0.3\textwidth} 
  \centering
  \small
  \caption{\footnotesize{Comparison of reward \\ computation in rule distillation}} 
  \label{tab:mcts_ablation} 
  \begin{tabular}{lcc}
    \toprule
    \textbf{Reward} & \textbf{$\mathcal{H}$} & \textbf{$JS$} \\ 
    \midrule
    PAR  & 3.791 & 0.127 \\ 
    Ours & 2.960 & 0.467 \\ 
    \bottomrule
  \end{tabular}
    \label{tab:mcts ablation}
\end{wraptable}
  We also analyzed the frequency of sub-rule occurrences, and the results in the Figure \ref{fig:PAR} indicate that sub-rules from PAR are more dispersed. These findings highlight the advantage of our reward computation in identifying stable and unified rules.

\begin{figure*}[h]
    \includegraphics[width=0.98\textwidth]{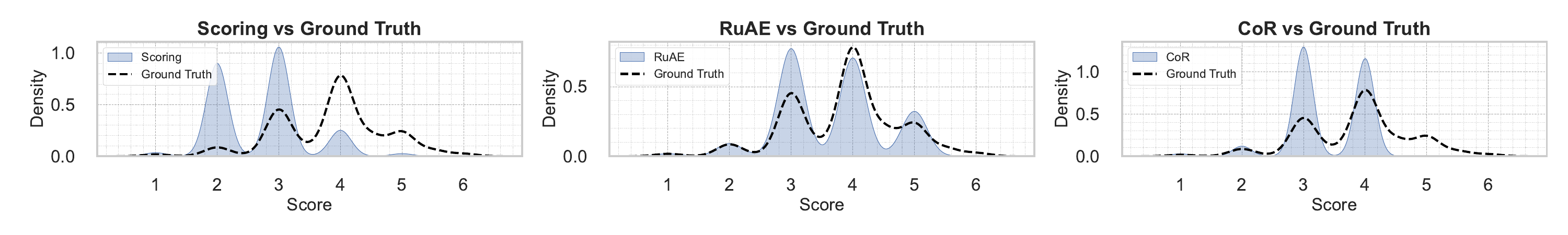} 
    \caption{Distributions (KDE) of predictions on ASAP}
    \label{fig:KDE}
\end{figure*}

\begin{figure*}[htbp] 
    \centering
    \includegraphics[width=1\textwidth, height=3.3cm]{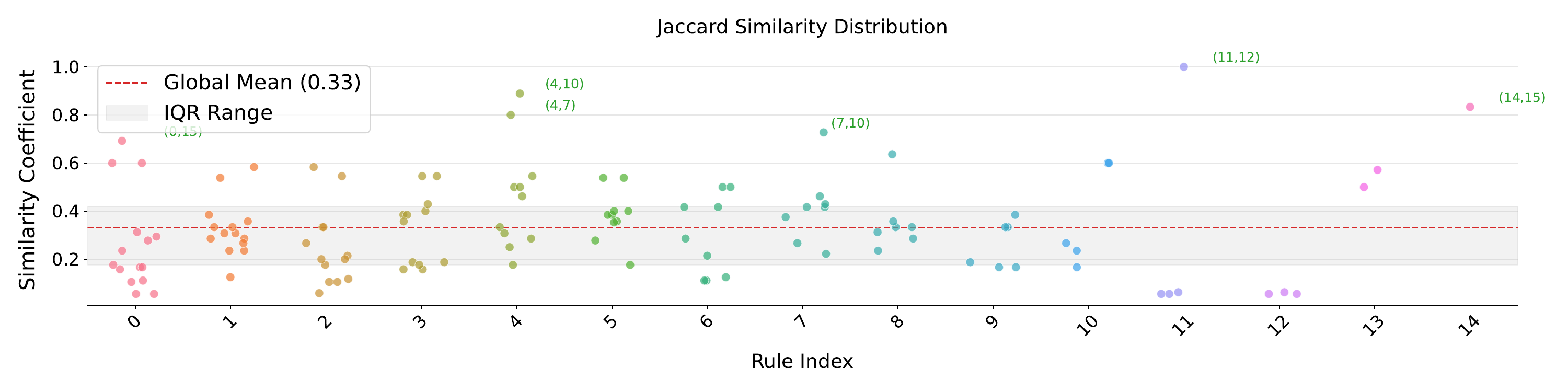} 
    \caption{Jaccard similarity among sub-rules} 
    \label{fig:roubst} 
\end{figure*}

\subsection{Analysis}
\textbf{Learned-rule Analysis}. We explore whether the rule distillation stage produces interpretable and practical scoring rules. Table \ref{tab:top5_combinations} presents the high-value rules (top 5) for four tasks, with "i" indicating the initial rubrics, "s" indicating stricter rubrics and "r" denoting a more lenient approach. Only aspects are shown, with detailed rubrics in Appendix \ref{appendix:Rules}.

For identifying relevant literature (Relish), application, findings are most critical, while method is less significant. This reflects the Relish’s biomedical focus, where these aspects align with disciplinary priorities. For rating prediction (Amazon), positive sentiment and satisfaction are key, aligning with common expectations. 
Essay scoring involves expert-defined evaluation aspects and guidelines (Details in Appendix \ref{appendix:Rules}), which allows us to evaluate the alignment between distilled rules and human-defined rules. Our evaluation metrics include precision (P), recall(R), Jaccard similarity($JS$), and performance above a random benchmark (LoR), along with a hypergeometric test to determine statistical significance ($p_{(HGT)}$).

\begin{table}[h!]
\small 
\centering
\caption{Sub-rules from high-value rules}
\label{tab:top5_combinations}
\begin{tabular}{ll}
    \toprule
    \textbf{Task} & \textbf{Set of high-value sub-rules} \\
    \midrule
    Relish & Applications(r,i), Finding(r); Goal(r),\\ & Domains(i), Idea(i)\\
    \midrule
    Amazon & Positive sentiment (s,i),\\
    & Satisfaction level (r,i) \\
    \midrule
    ASAP & Organization(i), Word choice(i,r) \\ 
 &  Idea\&Content(r), Sentence fluency (i),  \\
 &  Evidence support(i,r,s)\\
     \midrule
     SummEval & Coherence (r,i), Consistency (r,i) \\
     & Fluency (i), Relevance (r,s)\\
    \bottomrule
\end{tabular}
\end{table}

\begin{table}[h!]
\centering
\small
\begin{tabular}{lccccc}
\toprule
 & P& R & $JS$ & LoR & $p_{(HGT)}$ \\
\midrule
Value & 1.00 & 0.83 & 0.83 & 1.67 & 0.024 \\
\bottomrule
\end{tabular}
\caption{Evaluation metrics for alignment with human-defined rules.}
\label{tab:alignrule}
\end{table}
Our method achieved high precision and recall (in Tab \ref{tab:alignrule}), missing only the "Conventions" from human set. The LoR shows a 67\% improvement over random selection. Furthermore, hypergeometric test yields a p-value of 0.024, confirming that this high level of agreement is statistically significant.

\noindent\textbf{Score Alignment}. We evaluated the alignment between the score distributions of different methods and the ground truth, focusing on their performance on the ASAP dataset. Three methods, Scoring, RuAE, and CoR, were compared by using KDE plots. As shown in \ref{fig:KDE}. 
RuAE showed the closest alignment with the ground truth. 
Scoring method exhibited significant deviations, mostly predicting scores of 2 and 3 and underestimating the more common score of 4. CoR predictions were overly concentrated in the 3-4 score range, but its distribution was closer to the ground truth than Scoring. Therefore, CoR could effectively distinguish between poor, mediocre, and excellent essays, though some bias remained. Both CoR and RuAE showed a tendency to concentrate predictions around 3 and 4, but after reinforcement learning, RuAE significantly reduce the bias of CoR.

\noindent\textbf{Robust of sub-rule generation}. 
We examine the robustness of rule generation in rule distillation, specifically whether the initially generated sub-rules exhibit significant variation. We repeated the candidate rule generation process 15 times at a higher temperature (1.5) and calculated pairwise Jaccard similarity  among the generated rules. The results are presented in the figure below \ref{fig:roubst}. Most rules have similarity coefficients of 0.3 to 0.4, indicating high similarity among candidates, especially for rules 1 to 6, which closely resemble many others. This suggests limited variability when tasks and prerequisites are similar, the variability in rules is limited. Even if the wording differs, the underlying meaning remains comparable. This reflects the stability of action generation in MCTS.

\section{Conclusion}
We introduced a rule-guided approach for general LLM evaluators. We first use MCTS-based rule search to distill consistent scoring rules from data. Then, we enhanced LLM reasoning with these rules using two strategies: CoR and RuAE. Experiments on various tasks showed that our approach improves evaluation accuracy, especially for complex reasoning, and produces human-aligned evaluation rules. Our work demonstrates the potential of making LLMs better general evaluators.

\newpage
\section*{Limitations}
There are several limitations to our approach. First, it is more effective for tasks involving complex analysis and reasoning, while offering limited improvement for tasks that do not require strong reasoning abilities. Second, although the MCTS-based method greatly reduces search complexity, the overall computational cost remains high due to reliance on LLMs for score prediction and reward calculation. Furthermore, restricting the action space \(A_{\text{modify}}\) to either stricter or looser scoring criteria limits the extent of possible modifications. Third, for many evaluation tasks, a unified set of evaluation rules is not followed; instead, diversity in assessment is more important for these tasks. In such cases, our method may fail to deliver effective results. Finally, reinforcement learning training itself requires substantial computational resources.

\bibliography{anthology,custom}
\newpage
\newpage

\appendix

\section{Details of Experiment}
\label{appendix:details}

\subsection{Dataset}
We utilized four diverse datasets to evaluate the performance of our method: ASAP, Relish, Amazon, and SummEval. Each dataset was divided into training and test sets to ensure a robust evaluation and comparison. The distribution of these sets is presented in Table \ref{tab:dataset}. Except for the SummEval dataset, each of the other datasets contains 1200 test samples. In the phase of rule distillation, 200 samples were uesed.
\begin{itemize}
    \item \textbf{ASAP}: The Automated Student Assessment Prize (ASAP) dataset, sourced from Kaggle, is used for evaluating automated essay scoring systems.We selected over 3,500 samples from Prompt-1 and Prompt-2 out of the 8 available prompts in the dataset. The essays are scored on a scale from 1 to 6.  \url{https://www.kaggle.com/c/asap-aes/data}.
    \item \textbf{Relish}:The RElevant LIterature SearcH (Relish) dataset is a joint annotation effort involving more than 1500 scientists. The dataset uses a three-point labeling scheme: 2 for relevant, 1 for partially. relevant, and 0 for irrelevant.  A higher score indicates stronger relevance. \url{https://huggingface.co/datasets/allenai/scirepeval_test/viewer/relish}.
    \item \textbf{Amazon}  (US Amazon Reviews):This dataset consists of product reviews from Amazon, with the goal of predicting the 5-star rating for each review \citep{ni2019amzon}.
    \item \textbf{SummEval}: The SummEval dataset consists of 100 input source texts, each paired with 16 summary candidates generated by different language models.  Half of the data is used for training, and the other half for testing. The dataset is annotated across four aspects: coherency (CH), fluency (FLU), consistency (CON), and relevancy (RE). The quality of each generated text is represented by the average score across these four aspects.
    \url{https://huggingface.co/datasets/mteb/summeval}.
\end{itemize}

\begin{table}[ht]
\centering
\caption{Dataset Statistics.}
\label{tab:dataset}
\begin{tabular}{lccc}
\hline
\textbf{Dataset} & \textbf{Train} & \textbf{Test} & \textbf{Total} \\ \hline
ASAP             & 2382           & 1200          & 3582           \\
Relish           & 2398           & 1200          & 3598           \\
Amazon           & 1812           & 1200          & 3012           \\
SummEval         & 800            & 800           & 1600           \\ \hline
\end{tabular}
\end{table}

\subsection{Details of rule distillation}
In the phase of MCTS exploration, we configured specific parameters for each of the four task. These parameters include the maximum number of sub-rules to evaluate, the maximum number of aspects that can be explored when generating sub-rules, and the evaluation metric used to calculate the reward. The detailed settings for each task are presented in the table \ref{tab:mcts_settings}. For ASAP, we have known scoring aspects, including organization and word choice (for details, refer to Appendix \ref{appendix:Rules}). Therefore, when using the prompt template from Appendix \ref{appendix:Prompt}, we specify these six existing aspects for generating sub-rules while also allowing the exploration of additional potential scoring aspects. For SummEval, the specified aspects are coherency, fluency, consistency, and relevancy. For Relish, we prompt the model to consider scientific elements such as Goal and Method. For Amazon, no specific aspects are designated to allow for broader exploration. All CoT baselines are also prompted with the same evaluation aspects.

\begin{table}[h]
\centering
\caption{MCTS Settings for Each Task}
\label{tab:mcts_settings}
\small
\begin{tabular}{lccc}
\toprule
Task      & Max-Sub-Rules & Max-Aspects &  Metric \\
\midrule
ASAP      & 6             & 10          & QWK               \\
Amazon    & 3             & 7           & MSE               \\
Relish    & 4             & 8           & MAP               \\
SummEval  & 4             & 4           & $\rho$            \\
\bottomrule
\end{tabular}
\end{table}

\subsection{Details of Reinforcement Learning}
In the reinforcement learning phase, we perform fine-tuning based on the ms-SWIFT framework: \url{https://github.com/modelscope/ms-swift.git}. The training setup includes a maximum response length of 2048 tokens and a maximum prompt length of 2048 tokens. The training is conducted over a single epoch with a learning rate of 6e-7 and a temperature of 1.0. We utilize vLLM as the inference framework for rollout, with the `maximum length` set to 2048 and the precision configured as bfloat16.

\section{Results of Learned Rules and Analysis}
\label{appendix:Rules}

Table \ref{tab:rules} presents the candidate pool of sub-rules, the rules distilled, and the manually designed rules by human. The detailed rubrics are in: \url{https://lwsam.github.io/ASAP++/lrec2018.html} This comparison allows us to evaluate how well the algorithm's rule distillation aligns with human judgment. We employed the hypergeometric test to determine if the observed alignment between algorithm and human selections could occur by chance:
Null hypothesis ($H_0$): The algorithm randomly selects rules from the pool with no relationship to human choices
Alternative hypothesis ($H_1$): The algorithm's selections show significant alignment with human choices.
The hypergeometric test yielded a p-value of 0.0238, calculated as:
$P(X \geq 5) = \frac{C(6,5) \times C(4,0)}{C(10,5)} = \frac{6}{252} \approx 0.0238$

The Performance Above Random Benchmark calculation quantifies how much better the algorithm performed compared to random selection. We calculated this as: $(Actual\ matches - Expected\ matches) / Expected\ matches$. The expected number of matches if selection were random would be: $|\text{pred}| \times \frac{|\text{human}|}{|\text{pool}|} = 5 \times \frac{6}{10} = 3$ rules. Since the algorithm actually matched 5 rules, the performance improvement is: $(5-3)/3 = \frac{2}{3} = 66.7\%$. This indicates that the algorithm's rule selection performance exceeds random selection by 66.7\%, demonstrating substantial algorithmic learning of human-relevant evaluation criteria rather than chance alignment.

\begin{figure*}[t]
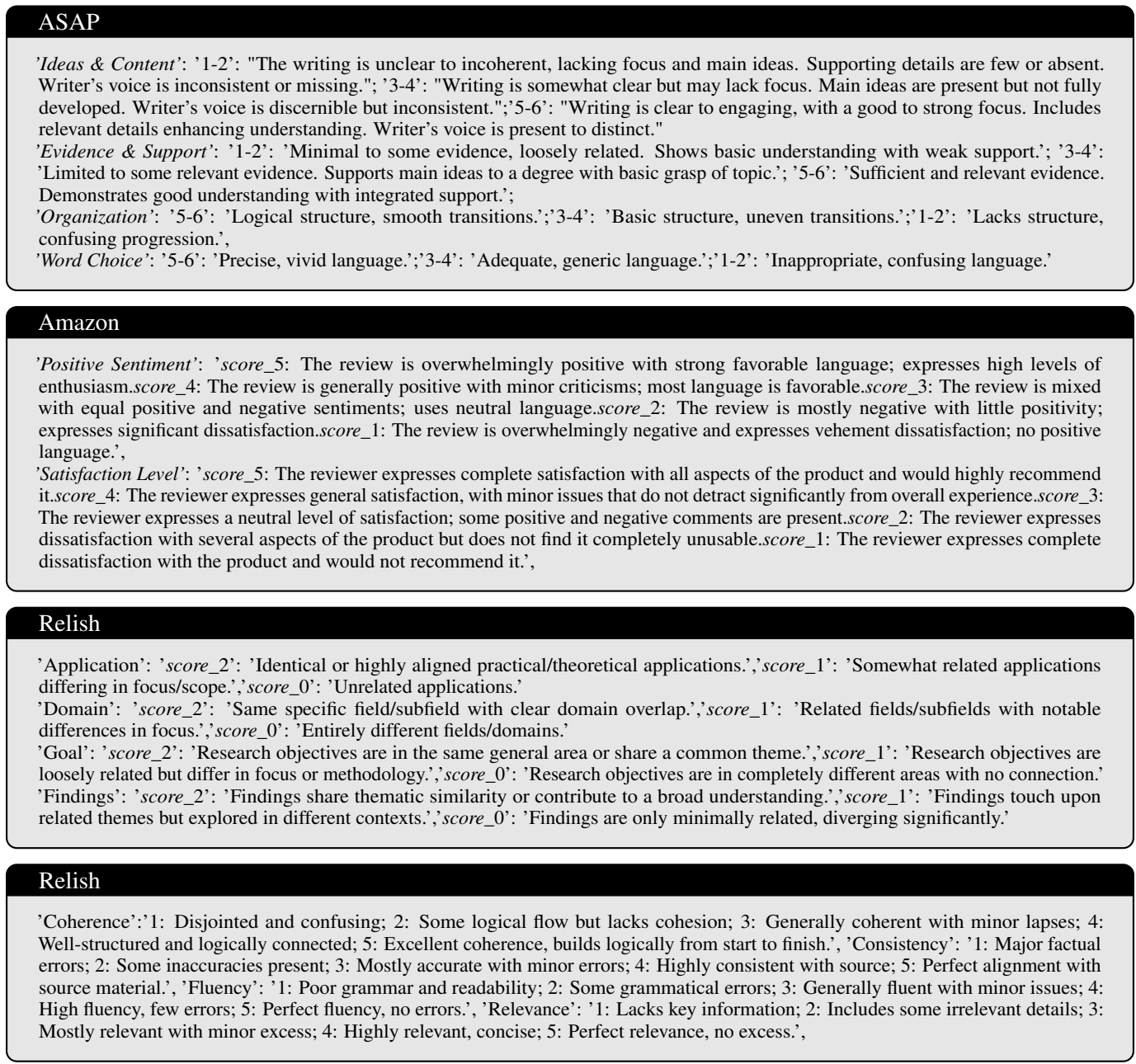


\begin{tcolorbox}[title = ASAP, colback=gray!20,
                  colframe=black,
                  width=18cm,
                  arc=1.5mm, auto outer arc,
                  boxrule=0.8pt,
                 ]
                 \small 
\textit{'Ideas \& Content'}: '1-2': "The writing is unclear to incoherent, lacking focus and main ideas. Supporting details are few or absent. Writer's voice is inconsistent or missing."; '3-4': "Writing is somewhat clear but may lack focus. Main ideas are present but not fully developed. Writer's voice is discernible but inconsistent.";'5-6': "Writing is clear to engaging, with a good to strong focus. Includes relevant details enhancing understanding. Writer's voice is present to distinct."

 \textit{'Evidence \& Support'}: '1-2': 'Minimal to some evidence, loosely related. Shows basic understanding with weak support.'; '3-4': 'Limited to some relevant evidence. Supports main ideas to a degree with basic grasp of topic.'; '5-6': 'Sufficient and relevant evidence. Demonstrates good understanding with integrated support.';

  \textit{'Organization'}: '5-6': 'Logical structure, smooth transitions.';'3-4': 'Basic structure, uneven transitions.';'1-2': 'Lacks structure, confusing progression.',
   
  \textit{'Word Choice'}: '5-6': 'Precise, vivid language.';'3-4': 'Adequate, generic language.';'1-2': 'Inappropriate, confusing language.'

\end{tcolorbox}

\begin{tcolorbox}[title = Amazon, colback=gray!20,
                  colframe=black,
                  width=18cm,
                  arc=1.5mm, auto outer arc,
                  boxrule=0.8pt,
                 ]
                 \small 
\textit{'Positive Sentiment'}: '\textit{score}\_5: The review is overwhelmingly positive with strong favorable language; expresses high levels of enthusiasm.\textit{score}\_4: The review is generally positive with minor criticisms; most language is favorable.\textit{score}\_3: The review is mixed with equal positive and negative sentiments; uses neutral language.\textit{score}\_2: The review is mostly negative with little positivity; expresses significant dissatisfaction.\textit{score}\_1: The review is overwhelmingly negative and expresses vehement dissatisfaction; no positive language.',

\textit{'Satisfaction Level'}: '\textit{score}\_5: The reviewer expresses complete satisfaction with all aspects of the product and would highly recommend it.\textit{score}\_4: The reviewer expresses general satisfaction, with minor issues that do not detract significantly from overall experience.\textit{score}\_3: The reviewer expresses a neutral level of satisfaction; some positive and negative comments are present.\textit{score}\_2: The reviewer expresses dissatisfaction with several aspects of the product but does not find it completely unusable.\textit{score}\_1: The reviewer expresses complete dissatisfaction with the product and would not recommend it.',

\end{tcolorbox}

\begin{tcolorbox}[title = Relish, colback=gray!20,
                  colframe=black,
                  width=18cm,
                  arc=1.5mm, auto outer arc,
                  boxrule=0.8pt,
                 ]
                 \small 
'Application': '\textit{score}\_2': 'Identical or highly aligned practical/theoretical applications.','\textit{score}\_1': 'Somewhat related applications differing in focus/scope.','\textit{score}\_0': 'Unrelated applications.'

 'Domain': '\textit{score}\_2': 'Same specific field/subfield with clear domain overlap.','\textit{score}\_1': 'Related fields/subfields with notable differences in focus.','\textit{score}\_0': 'Entirely different fields/domains.'

'Goal': '\textit{score}\_2': 'Research objectives are in the same general area or share a common theme.','\textit{score}\_1': 'Research objectives are loosely related but differ in focus or methodology.','\textit{score}\_0': 'Research objectives are in completely different areas with no connection.'

 'Findings': '\textit{score}\_2': 'Findings share thematic similarity or contribute to a broad understanding.','\textit{score}\_1': 'Findings touch upon related themes but explored in different contexts.','\textit{score}\_0': 'Findings are only minimally related, diverging significantly.'

\end{tcolorbox}

\begin{tcolorbox}[title = Relish, colback=gray!20,
                  colframe=black,
                  width=18cm,
                  arc=1.5mm, auto outer arc,
                  boxrule=0.8pt,
                 ]
                 \small 
'Coherence':'1: Disjointed and confusing; 2: Some logical flow but lacks cohesion; 3: Generally coherent with minor lapses; 4: Well-structured and logically connected; 5: Excellent coherence, builds logically from start to finish.',
 'Consistency':  '1: Major factual errors; 2: Some inaccuracies present; 3: Mostly accurate with minor errors; 4: Highly consistent with source; 5: Perfect alignment with source material.',
 'Fluency': '1: Poor grammar and readability; 2: Some grammatical errors; 3: Generally fluent with minor issues; 4: High fluency, few errors; 5: Perfect fluency, no errors.',
 'Relevance':  '1: Lacks key information; 2: Includes some irrelevant details; 3: Mostly relevant with minor excess; 4: Highly relevant, concise; 5: Perfect relevance, no excess.',
\end{tcolorbox}
\caption{The learned rule from rule distillation phase.}
\label{fig:rules learned}
\end{figure*}

Figure \ref{fig:rules learned} shows the rules learned during the MCTS phase, which cover both scoring aspects and specific criteria. Due to space limitations, we only present the rules with the highest average rewards.

\begin{table*}[htbp]
\centering
\caption{Rule sets for evaluation}
\begin{tabularx}{\textwidth}{lX}
\toprule
\textbf{Type} & \textbf{Rules} \\
\midrule
Candidate Pool & \makecell[l]{
"Ideas \& Content", "Organization", "Word Choice", "Sentence Fluency", "Conventions",\\
"Originality \& Creativity", "Evidence \& Support", "Audience Awareness",\\
"Clarity \& Coherence", "Engagement \& Impact"
} \\
Human-Defined & \makecell[l]{
"Ideas \& Content", "Organization", "Word Choice",
"Sentence Fluency",\\ "Conventions", "Evidence \& Support"
} \\
Distillation & \makecell[l]{
"Evidence \& Support", "Organization", "Word Choice",\\
"Sentence Fluency", "Ideas \& Content"
} \\
\bottomrule
\end{tabularx}
\label{tab:rules}
\end{table*}

\section{Prompt}
\label{appendix:Prompt}

The Sub-rules Generation Prompt is shown in Figure \ref{fig:subprompt}, while the prompts related to CoR and RL data construction are presented in Figure \ref{fig:corprompt}.

\begin{figure*}[t]
\begin{tcolorbox}[title = Sub-rule Generation, colback=gray!20,
                  colframe=black,
                  width=18cm,
                  arc=1.5mm, auto outer arc,
                  boxrule=0.8pt,
                 ]
                 \small 
Generate \textit{{num}} assessment dimensions beyond the existing ones:\\
\textit{{existing aspects}}:

The dimension should be: \\- As independent as possible; - Objectively measurable;- Add meaningful perspective to the assessment.\\
For each scoring dimension, develop detailed scoring guidelines. Each guideline should be described in several intervals that collectively cover the entire scoring range, with a total description limit of 60 words. The guideline should be Clear, specific, and Highly instructive,explaining what qualifies for the score. \\
Please parse the "Assessment\_Dimensions";"Scoring\_Guideline "; output them in JSON format.\\
Example JSON:
{{\\
    "Assessment\_Dimensions": ["Sentiment";"Aspect2";"Aspect3"],\\
    "Scoring\_Guideline ":["1-2: The review is mostly negative with little positivity; expresses significant dissatisfaction; 3-4: Somewhat negative, with minimal positive aspects, ...; Score 5 and above - Generally positive, few negative elements.";"Guideline for Aspect2"; ...]
}}
\end{tcolorbox}

\begin{tcolorbox}[title = Rubrics modification, colback=gray!20,
                  colframe=black,
                  width=18cm,
                  arc=1.5mm, auto outer arc,
                  boxrule=0.8pt,
                 ]
                 \small 
\#\# Make the scoring criteria more \textbf{lenient}.
This is the scoring dimensions and guidelines. Your task is to rewrite each Scoring\_Guideline to be much more lenient and significantly different from the original. Under the new guideline, the same text should receive a higher score.\\
Please maintainthe same JSON output format as the original.\\
\#\# scoring dimensions and guidelines:
\textit{{guidelines}}
\\

\#\# Make the scoring criteria more \textbf{stricter}.
This is the scoring dimensions and guidelines. Your task is to rewrite each Scoring\_Guideline to be much more stricter and significantly different from the original. Under the new guideline, the same text should receive a higher score.\\
Please maintainthe same JSON output format as the original.\\
\#\# scoring dimensions and guidelines:
\textit{{guidelines}}

\end{tcolorbox}
\caption{Sub-rule generation Prompt}
\label{fig:subprompt}
\end{figure*}

\begin{figure*}[t]

\begin{tcolorbox}[title = Chain-of-Rule Prompt, colback=yellow!20,
                  colframe=gray,
                  width=18cm,
                  arc=1.5mm, auto outer arc,
                  boxrule=0.8pt,
                 ]
                 \small 
You are tasked with evaluating the text based on the given Scoring Criteria:

\textcolor{blue}{\textbf{Aspect-1}}:\textit{Rubric-1}\\
\textcolor{blue}{\textbf{Aspect-2}}:\textit{Rubric-2} \\

\#\# Texts to be evaluated\\
Text-1: Essay1: "Dear Local Newspaper, @CAPS1 are so many ....\\

For each criterion, provide a brief analysis and assign scores. Then, provide a comprehensive score upon them. The final score ranges from a to b.

\end{tcolorbox}

\begin{tcolorbox}[title = Rule-guided Prompt, colback=yellow!20,
                  colframe=gray,
                  width=18cm,
                  arc=1.5mm, auto outer arc,
                  boxrule=0.8pt,
                 ]
                 \small 
First, you should choose no more than {\textit{num}} scoring criteria from the given candidate Scoring Criteria that you think are important for the task. Then, assign weights to these chosen aspect.
You are tasked with evaluating a pair of texts based on the given Scoring Criteria you have selected.

\textcolor{blue}{\textbf{Aspect-1}}:\textit{Rubric-1}\\
\textcolor{blue}{\textbf{Aspect-2}}:\textit{Rubric-2} \\
\textcolor{blue}{\textbf{Aspect-3}}:\textit{Rubric-3}\\

\#\# Texts to be evaluated\\
Text-1: Essay1: "Dear Local Newspaper, @CAPS1 are so many ....\\
Text-2: The use of computer benefits society due to the possitive affect ...

For each criterion, provide a brief analysis and assign scores. 
Then, provide a comprehensive score upon them. The final score ranges from a to b.\\

\#\# Output Format Requirements\\
Selected Aspects: <Aspects and weights>. The key evaluation aspects you selected are enclosed within <Aspect> </Aspect> tags. For example, <Aspect>Content, Goal</Aspect>, with weights: \verb|\weighted{x,y}|.\\
Analysis: <Compare and evaluate different texts based on given Criteria from multiple aspects you've selected>. The Analysis are enclosed within <Analysis> </Analysis> tags.\\
Scores: <Based on analysis and scores of this pair of texts across various aspects and considering the weights of aspects, synthesize an overall score. The final scores are placed in \verb|\box{x,y}|.
\end{tcolorbox}
\caption{Chain-of-Rule Prompt and Rule-guided Prompt}
\label{fig:corprompt}
\end{figure*}

\end{document}